\documentclass[sigconf, nonacm, natbib=true]{acmart}

\PassOptionsToPackage{icons=false}{doclicense}

\usepackage{booktabs}     % 提供 \toprule, \midrule, \bottomrule
\usepackage{multirow}     % 提供 \multirow
\usepackage{amsmath}      % 提供数学公式支持
\usepackage{graphicx}     % 提供 \resizebox
\usepackage{subcaption}
\usepackage{tabularx}     % 提供 tabularx 环境和 X 列类型
\usepackage[table]{xcolor} % 提供 \rowcolor (注意必须带 [table] 参数)
\newcommand{\specialcell}[2][c]{%
  \begin{tabular}[#1]{@{}c@{}}#2\end{tabular}}
\newcolumntype{C}{>{\centering\arraybackslash}X}
\AtBeginDocument{%
  }
\renewcommand\footnotetextcopyrightpermission[1]{}
\setcopyright{acmlicensed}
\copyrightyear{2018}
\acmYear{2018}
\acmDOI{XXXXXXX.XXXXXXX}
\acmConference[Conference acronym 'XX]{Make sure to enter the correct
  conference title from your rights confirmation email}{June 03--05,
  2018}{Woodstock, NY}
\acmISBN{978-1-4503-XXXX-X/2018/06}
\begin{document}

\title{SpatialFusion: Endowing Unified Image Generation with Intrinsic 3D Geometric Awareness}

\author{Haiyi Qiu}
\authornote{Equal contribution}
\affiliation{%
  \institution{Zhejiang University}
  \city{Hangzhou}
  \country{China}
}
\email{haiyiqiu@zju.edu.cn}

\author{Kaihang Pan}
\authornotemark[1]
\affiliation{%
  \institution{Zhejiang University}
  \city{Hangzhou}
  \country{China}
}
\email{kaihangpan@zju.edu.cn}

\author{Jiacheng Li}
\affiliation{%
  \institution{HiThink Research}
  \city{Hangzhou}
  \country{China}
}
\email{lijiacheng2@myhexin.com}

\author{Juncheng Li}
\authornote{Corresponding author}
\affiliation{%
  \institution{Zhejiang University}
  \city{Hangzhou}
  \country{China}
}
\email{junchengli@zju.edu.cn}

\author{Siliang Tang}
\affiliation{%
  \institution{Zhejiang University}
  \city{Hangzhou}
  \country{China}
}
\email{siliang@zju.edu.cn}

\author{Yueting Zhuang}
\affiliation{%
  \institution{Zhejiang University}
  \city{Hangzhou}
  \country{China}
}
\email{yzhuang@zju.edu.cn}

% \renewcommand{\shortauthors}{Qiu et al.}

\begin{abstract}
Recent unified image generation models have achieved remarkable success by employing MLLMs for semantic understanding and diffusion backbones for image generation. However, these models remain fundamentally limited in spatially-aware tasks due to a lack of intrinsic spatial understanding and the absence of explicit geometric guidance during generation. In this paper, we propose \textbf{SpatialFusion}, a novel framework that internalizes 3D geometric awareness into unified image generation models. Specifically, we first employ a Mixture-of-Transformers (MoT) architecture to augment the MLLM with a parallel spatial transformer to enhance 3D geometric modeling capability. By sharing self-attention with the MLLM, the spatial transformer learns to derive metric-depth maps of target images from rich semantic contexts. These explicit geometric scaffolds are then injected into the diffusion backbone through a specialized depth adapter, providing precise spatial constraints for spatially-coherent image generation. Through a progressive two-stage training strategy, \textbf{SpatialFusion} significantly enhances performance on spatially-aware benchmarks, notably outperforming leading models such as GPT-4o. Additionally, it achieves generalized performance gains across both text-to-image generation and image editing scenarios, all while maintaining negligible inference overhead. 
\end{abstract}

\begin{CCSXML}
<ccs2012>
   <concept>
       <concept_id>10010147.10010178.10010224.10010245</concept_id>
       <concept_desc>Computing methodologies~Computer vision problems</concept_desc>
       <concept_significance>500</concept_significance>
       </concept>
 </ccs2012>
\end{CCSXML}

\ccsdesc[500]{Computing methodologies~Computer vision problems}
\keywords{Unified Image Generation, 3D Geometric Awareness}


\maketitle

\section{Introduction}
Unified image generation models have made remarkable progress in recent years  \cite{deng2025emerging, wu2025omnigen2, xie2024show, lin2025uniworld, pan2025generative, pan2025janus}. Following the principle of understanding-reinforced generation \cite{wu2025omnigen2,pan2026omniweaving}, these approaches typically employ a Multimodal Large Language Model (MLLM) \cite{wu2025qwen, bai2025qwen2} as the understanding component to capture rich semantic guidance from text prompts, and a diffusion backbone \cite{peebles2023scalable, pan2025focusdiff, hu2025d} as the generation component to model conditional noise‑to‑image distribution, thereby enabling impressive capabilities for producing prompt‑consistent and realistic images.

\begin{figure}[t]
  \centering
  \includegraphics[width=\linewidth]{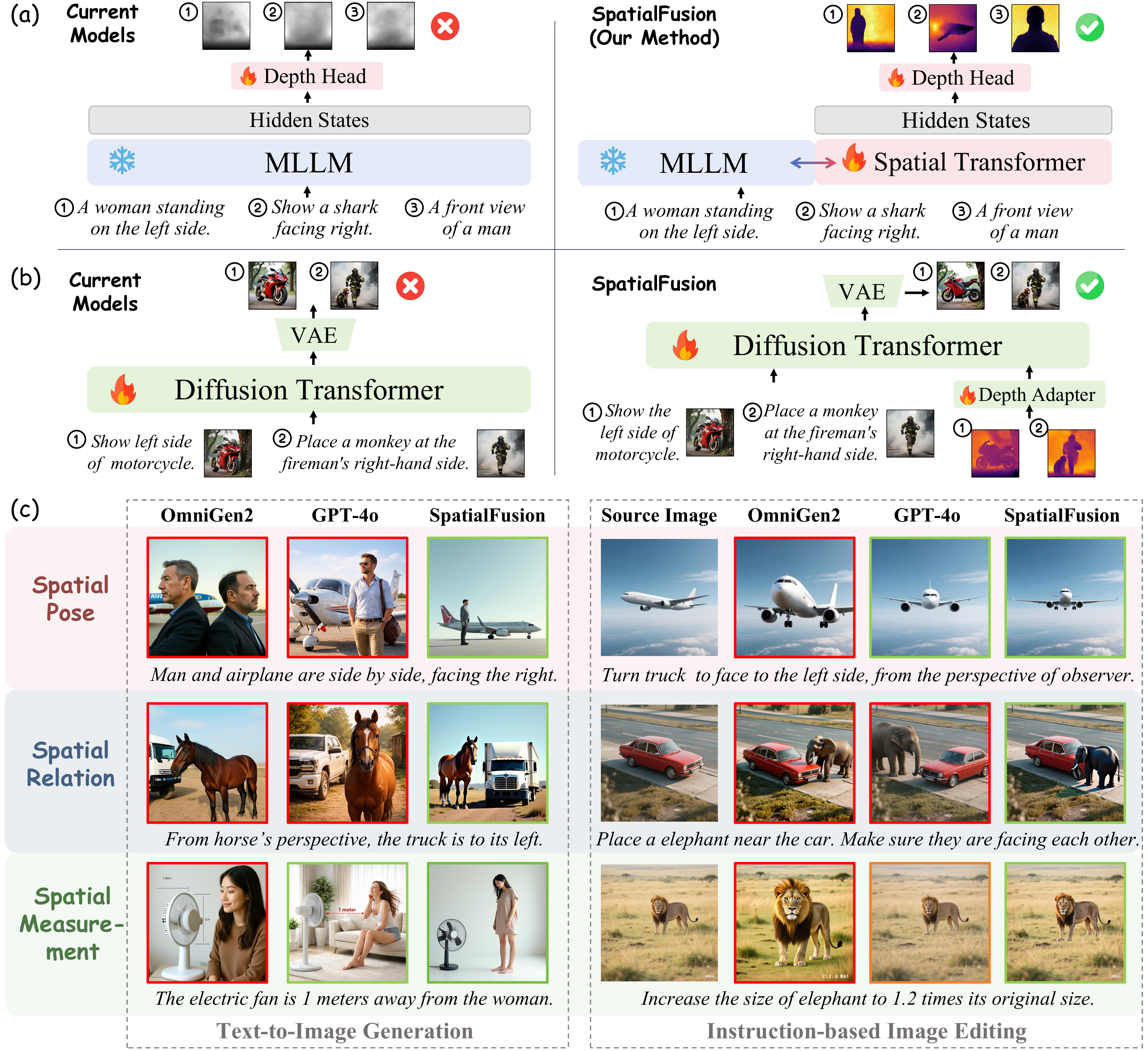}
  \caption{(a) MLLMs produce geometry-deficient hidden states that fail to be probed for depth maps;  (b) Diffusion models need explicit 3D guidance to achieve spatially-aware synthesis. (c) Limitations in spatially-aware generation tasks and enhanced performance of our method \textbf{SpatialFusion}.}
  \label{fig:first}
  \Description{}
\end{figure}

However, despite these advances, recent studies \cite{huang2025smartspatial, wang2025genspace, wang2026everything, su2026generation, pan2025wiseedit, hu2026reinforcement} reveal that even the most advanced models \textbf{exhibit fundamental limitations in spatially-aware tasks}, struggling with challenges such as understanding object poses, reasoning about spatial relations, and adhering to metric measurements, as shown in Fig. ~\ref{fig:first}(c). We argue that this limitation stems from the absence of intrinsic 3D geometric awareness in current models, where geometry-deficient representations in the understanding stage propagate into geometry-unconstrained synthesis in generation. \textbf{1) Geometry-deficient representations in understanding}. We conduct a probing experiment by fine-tuning a depth prediction head on the frozen MLLM’s hidden states in Fig. \ref{fig:first}(a). Results show that without specialized spatial modules, the MLLM's internal representations fail to generate any semantically-aligned depth maps. Empirical studies \cite{wu2025spatial, zhang2026gap, bigverdi2025perception, park2024geometry} further reveal that MLLM features correlate strongly with categorical concepts but weakly encode geometric attributes, thus rendering them unable to provide a reliable internal 3D scaffold. \textbf{2) Geometry-unconstrained synthesis in generation}. When the MLLM fails to provide geometry‑aware conditions, the diffusion backbone is forced to rely solely on high-level semantic embeddings.  We examine this generative behavior in Fig. \ref{fig:first}(b), finding that since the diffusion process is fundamentally an iterative denoising procedure without a natural structural prior \cite{ho2020denoising,saharia2022photorealistic,rombach2022high}, semantic embeddings are insufficient to anchor precise spatial arrangements. In contrast, when guided by dense geometric signals (e.g., depth maps), the model tends to produce synthesize images that demand coherent spatial structures and layouts. 

% This limitation stems from the absence of intrinsic 3D geometric perception and generation in existing models. \textbf{First, the understanding component lacks intrinsic capabilities for 3D geometric modeling}. Having been predominantly pre-trained on web-scale 2D image-text pairs for semantic alignment, MLLMs are inherently semantics-centric rather than geometry-aware. Empirical probing studies further reveal that while MLLM-derived features correlate strongly with categorical taxonomies, they exhibit weak encoding of structural and spatial properties. Consequently, these models struggle to infer or synthesize reliable 3D geometric structures from purely textual descriptions. \textbf{Second, the generation component lacks effective spatial guidance for image synthesis}. Relying solely on high-level semantic embeddings, diffusion backbones excel at modeling pixel-level distributions to conceptualize \textit{what} to generate, but possess little control over \textit{where} and \textit{how} objects should be situated within a coherent 3D space. Without explicit geometric priors such as metric-depth maps to constrain the sampling process, these models struggle to adhere to precise spatial requirements, including object pose, viewpoint, spatial relationships, and so on. \textbf{Therefore}, an ideal paradigm should internalize 3D geometric capabilities to address the deficiencies in both understanding and generation.

Previous works have made various attempts to provide geometric guidance. Layout-to-image methods \cite{li2023gligen, zheng2023layoutdiffusion, he2021context} incorporate 2D bounding boxes or segmentation maps for precise planar placement; however, their purely 2D nature precludes them from addressing 3D geometric tasks. Other approaches \cite{zhang2023adding, mou2024t2i, qin2023unicontrol} inject external 3D cues as conditioning signals, but their reliance on auxiliary information unavailable in text-only scenarios limits the applicability. More importantly, both lines of work function merely as conditional executors, failing to achieve the inherent geometric awareness.

Recent advances in 3D foundation models \cite{deng2025best,xu2026towards, wang2025vggt,shen2025fastvggt} offer a promising direction to address these challenges. In particular, models such as Visual Geometry Grounded Transformer (VGGT) \cite{wang2025vggt} demonstrate that rich 3D geometric information can be effectively learned within a unified Transformer architecture, suggesting that 2D visual semantics and 3D geometry can be formulated within a shared representational framework. Building on this insight, a pivotal question arises: \textit{can we extend this unified framework to generative models, enabling 2D semantics to derive 3D structural scaffold and, in turn, guide 2D image synthesis?}

% Strikingly, recent advances in both 3D vision and multi-modal architectures provide possibility to realize these goals. On one hand, the success of VGGT demonstrates that 3D geometric information—traditionally obtained through complex optimization—can now be efficiently extracted using a standard Transformer in a single feed-forward pass, making rich spatial cues readily accessible. On the other hand, the Mixture-of-Transformers (MoT) paradigm offers a principled framework for integrating multiple modalities within a unified architecture, allowing dedicated pathways to process distinct information streams while maintaining cross-modal interaction through shared attention mechanisms.

In this paper, we propose \textbf{SpatialFusion}, a novel framework that endows unified image generation models with intrinsic 3D geometric awareness. Specifically, it follows two synergistic mechanisms: \textbf{1) Semantics-Guided Geometric Derivation}: We introduce a parallel spatial transformer that operates alongside the MLLM for spatial understanding. Following the Mixture-of-Transformers (MoT) architecture \cite{shi2024lmfusion,liang2024mixture}, the spatial transformer maintains modality-specific parameters to specialize in 3D geometric modeling, while sharing self-attention layers with the MLLM to interact with semantic contexts. Guided by the multimodal semantics, the primary objective of the transfromer is to derive explicit 3D structures (i.e. metric-depth maps) of the final synthesized image, thereby providing  geometric guidance to steer the subsequent generative process.
\textbf{2) Geometry-Constrained Image Synthesis}: To actualize spatial guidance during generation, a dedicated depth adapter is designed to align the derived metric-depth maps with the diffusion model’s latent space, facilitating their fusion via element-wise addition before multiple denoising steps. This mechanism provides rigorous structural grounding that is absent in conventional semantics-only conditioning. By embedding these explicit spatial priors, the framework empowers the generative process to reason about intricate spatial configurations, ensuring that the synthesized imagery strictly adheres to 3D poses, spatial relations, and metric scales.

To realize \textbf{SpatialFusion}, we employ a progressive two-stage training strategy. First, we leverage the 3D foundation model VGGT as a geometric teacher, supervising the spatial transformer to derive accurate metric-depth maps for target images. Second, we jointly optimize this pretrained transformer with the diffusion backbone, using the derived geometric scaffolds to steer the denoising process. Extensive experiments demonstrate that \textbf{SpatialFusion} significantly enhances model performance on the spatially-aware generation benchmark GenSpace \cite{wang2025genspace}, notably outperforming leading models including GPT-4o \cite{openai2025gpt4o}. Our approach also achieves broad performance gains across general T2I and editing scenarios, proving that intrinsic geometric awareness effectively elevates the model's broader generative and reasoning capabilities.

Overall, our contributions are as follows:
\begin{itemize}
    \item We propose \textbf{SpatialFusion}, a novel framework to internalizes 3D geometric awareness within generative models, aiming to resolve the absence of intrinsic spatial reasoning and explicit geometric constraints in 2D-centric generation.
    \item We introduce a synergistic mechanism, where semantics context drives the derivation of target 3D structures, which in turn act as explicit geometric guidance for diffusion-based 2D image synthesis.
    \item Extensive experiments demonstrate the superior performance of \textbf{SpatialFusion} on spatially-aware and standard benchmarks, achieving broad performance gains while maintaining negligible inference overhead.
\end{itemize}

\section{Related Work}

\subsection{Unified Image Generation Model}
Unified image generation models \cite{deng2025emerging, wu2025omnigen2, xie2024show, lin2025uniworld} have made remarkable progress in recent years, converging toward a common insight: that understanding can facilitate generation.
These models generally leverage powerful MLLMs as semantic condition encoders to guide the diffusion process, achieving strong performance in prompt alignment and visual realism. However, they operate solely in 2D space without intrinsic 3D geometric awareness, making them incapable of handling spatially-aware generation tasks that require object pose understanding, spatial relation reasoning, or metric measurement adherence \cite{huang2025smartspatial, wang2025genspace, wang2026everything, su2026generation} . In contrast, \textbf{SpatialFusion} addresses this limitation by internalizing 3D geometric awareness into the unified generation models. By enabling semantics-guided geometric derivation and geometry-constrained image synthesis, our method significantly improves spatial understanding and generation quality in spatially-aware tasks.

\subsection{Unified Geometry Foundation Model}
Parallel to advances in 2D generation \cite{hu2025towards,hu2025asynchronous}, the 3D vision community has witnessed the emergence of unified geometry foundation models \cite{deng2025best,xu2026towards, wang2025vggt,shen2025fastvggt}. A representative example is VGGT \cite{shen2025fastvggt}, which can directly infer all key 3D attributes of a scene (e.g., camera parameters, depth maps) from input views in a single forward pass. Notably, VGGT is built on a fairly standard large Transformer architecture, following the same mold as large language models and vision backbones, which opens the door to seamless integration into other modalities and tasks. Motivated by this, \textbf{SpatialFusion} leverages the architectural homogeneity of these Transformer-based models, internalizing 3D awareness into unified image generation models as a native capability rather than an external constraint.

\section{Method}
In this section, we introduce \textbf{SpatialFusion}, a unified framework that internalizes 3D geometric awareness into unified image generation models. We detail the model architecture, training tasks, and training strategy in Sec.~\ref{sec:3.1}, \ref{sec:3.2}, and ~\ref{sec:3.3}, respectively. More Details are given in Appendix A.

\begin{figure*}[t]
  \centering
  \includegraphics[width=0.95\textwidth]{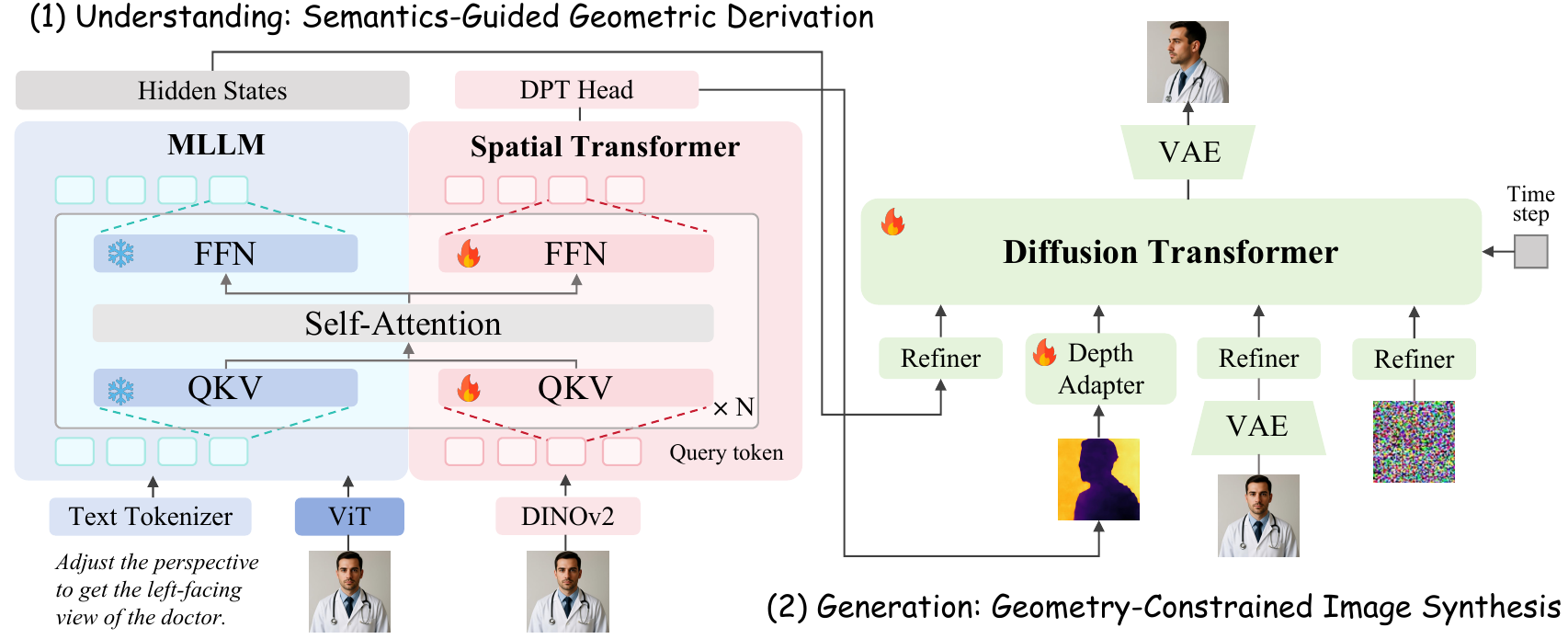}
  \caption{Overview of \textbf{SpatialFusion}. The framework internalizes 3D awareness via: (1) Semantics-Guided Geometric Derivation, where MoT aligns semantic guidance with metric-depth maps; and (2) Geometry-Constrained Image Synthesis, where derived geometric scaffolds steer the DiT toward spatially-controllable generation.}
  \label{fig:arch}
  \Description{}
\end{figure*}

\subsection{Model Architecture}
\label{sec:3.1}

As illustrated in Fig.~\ref{fig:arch}, \textbf{SpatialFusion} synergizes four core components: a Multimodal Large Language Model (MLLM) for semantic context parsing, a parallel Spatial Transformer for 3D geometric modeling, a Variational Autoencoder (VAE) for latent space compression, and a Diffusion Transformer (DiT) for high-fidelity image synthesis. By internalizing 3D awareness through the following two synergistic mechanisms, our architecture achieves superior spatial-semantic coherence and fidelity:

\textbf{(1) Semantics-Guided Geometric Derivation:} 
To address the lack of explicit geometric grounding in previous models, we introduce a parallel Spatial Transformer alongside the MLLM under a Mixture-of-Transformers (MoT) paradigm. The $L$-layer Spatial Transformer interacts with the $M$-layer MLLM at selected layers, thereby deriving structural representations from semantic context.
Specifically, we initialize a set of learnable query tokens, which are dynamically augmented with DINOv2 features from the input image, to form the initial geometric states $\mathbf{H}_{geo}^{(0)}$. At each interactive MoT layer $i$ (with $i = 1, \dots, L$), $\mathbf{H}_{geo}^{(i)}$ and $\mathbf{H}_{sem}^{(j)}$ (where $j = \lfloor i \cdot M/L \rfloor$, $M > L$) are first converted into their respective queries, keys, and values via separate projection matrices:
\begin{equation}
\mathbf{Q}_{geo}^{(i)}, \mathbf{K}_{geo}^{(i)}, \mathbf{V}_{geo}^{(i)} = \mathbf{QKV}_{geo}(\mathbf{H}_{geo}^{(i)}),
\end{equation}
\begin{equation}
\mathbf{Q}_{sem}^{(j)}, \mathbf{K}_{sem}^{(j)}, \mathbf{V}_{sem}^{(j)} = \mathbf{QKV}_{sem}(\mathbf{H}_{sem}^{(j)}).
\end{equation}
We then enable semantics-guided attention by concatenating the keys and values from both pathways into unified sequences, allowing geometric queries to build 3D geometric representations grounded on the MLLM's rich semantic context:
\begin{equation}
\mathbf{Attn}_{geo}^{(i)} = \text{softmax}\left( \frac{\mathbf{Q}_{geo}^{(i)} [\mathbf{K}_{sem}^{(j)} \circ \mathbf{K}_{geo}^{(i)}]^T}{\sqrt{d}} \right) [\mathbf{V}_{sem}^{(j)} \circ \mathbf{V}_{geo}^{(i)}],
\end{equation}
where $\circ$ denotes concatenation. $\mathbf{Attn}_{geo}^{(i)}$ is then passed through a geometric-specific output projection and a feed-forward network to produce the updated geometric states $\mathbf{H}_{geo}^{(i+1)}$. Finally, the geometric-aware last-layer states $\mathbf{H}_{geo}^{(L)}$ are fed into a DPT head \cite{ranftl2021vision} to decode a dense metric-depth map $\mathbf{D} \in \mathbb{R}^{H \times W \times 1}$.

\textbf{(2) Geometry-Constrained Image Synthesis:} 
To provide explicit spatial guidance, we leverage the derived 3D scaffolds $\mathbf{D}$ to constrain diffusion process. Specifically, the  metric-depth map $\mathbf{D}$ is first processed by a dedicated Depth Adapter $\mathcal{E}_{\phi}$ to align it with diffusion latent space: $F_{depth} = \mathcal{E}_{\phi}(\mathbf{D}) \in \mathbb{R}^{h \times w \times c}$ (where $h=H/8, w=W/8$). These features are then fused with the VAE-encoded noisy latents $z_t$ via element-wise addition:
\begin{equation}
\hat{z}_t = z_t \oplus F_{depth}, 
\end{equation}
This fusion occurs before latent patchification, embedding 3D constraints at the foundational spatial level. The DiT backbone then processes the geometry-augmented noisy latents alongside the MLLM's last hidden states $\mathbf{H}_{sem}^{(M)}$ as a unified sequence, predicting noise via $\epsilon\theta(\hat{z}t, t, \mathbf{H}_{sem}^{(M)})$ at each denoising step. This enables spatially-aware denoising that respects both semantics and geometry, yielding synthesized outputs that are not only semantically aligned with the text prompts but also strictly adhere to the underlying geometric structures.
\subsection{Training Tasks}
\label{sec:3.2}
While conventional text-to-image paired data provides essential semantic supervision, it falls short in supporting complex spatially-aware synthesis and consistent editing. 
To address this limitation, we construct a large-scale geometry-augmented dataset designed to facilitate rigorous geometric awareness for unified image generation. Our training tasks are structured around three primary competencies, each encompassing a wide variety of task formats.

\textbf{Foundational Image Generation Tasks:} This category establishes the baseline capabilities for unified image generation. It comprises two core tasks: (a) Standard text-to-image generation, which ensures robust alignment between textual prompts and visual semantics; and (b) Instruction-guided image-to-image editing, which trains the model to perform local and global visual modifications based on natural language commands, serving as a prerequisite for more sophisticated generative capabilities.

\textbf{Spatially-Aware Generation Tasks:} 
Translating semantic intent into precise spatial arrangements requires an explicit understanding of 3D structures. We formulate tasks that place strong emphasis on geometric conditions, with representative examples spanning three progressive levels of spatial intelligence: (a) Spatial Pose synthesis, where prompts dictate the 3D position and orientation of objects and cameras; (b) Spatial Relation reasoning, which requires the model to accurately render multi-object spatial layouts and relative positioning across different visual perspectives; and (c) Spatial Measurement adherence, demanding precise controllability over quantitative spatial details, such as object sizes and intervals.

\textbf{Geometry-Consistent Editing Tasks:} High-quality image editing demands that modifications blend naturally without disrupting the underlying 3D scene structure. Accordingly, We construct a dataset covering a variety of spatially-aware editing operations, including: (a) Geometry-grounded object addition, which requires placing new elements at plausible depths consistent with the perspective; (b) Structural-aware object removal,  which demands the model to inpaint the background while adhering to the original geometric scaffold; (c) Subject replacement, which ensures that the new entity conforms precisely to the spatial footprint of the original object; and other tasks involving spatial constraints. 

Overall, we curate a million-scale training dataset comprising web-sourced and synthetic samples. More details is in Appendix B.
\subsection{Training Strategy}
\label{sec:3.3}
Leveraging the proposed architecture and training tasks, we formulate a progressive, two-stage training paradigm, endowing unified image generation models with 3D geometric awareness.

\textbf{Stage 1: Geometric-Aware Pre-training}. 
In this stage, we freeze the MLLM and diffusion backbone, focusing exclusively on pre-training the Spatial Transformer and its learnable query tokens to derive geometric representations. We curate training subsets from the benchmarks described in Sec. \ref{sec:3.2}, prioritizing tasks that demand rigorous structural adherence. To provide dense geometric supervision, we leverage a frozen 3D foundation model (VGGT) as a teacher to extract pseudo-ground-truth metric-depth maps $D_{gt}$ directly from the target ground-truth images. Specifically, the learnable query tokens are processed by modality-specific layers and shared attention blocks, with the resulting hidden states fed into VGGT’s frozen DPT head for metric-depth prediction $D_{pred}$. To supervise this semantics-guided geometric derivation, we optimize the Spatial Transformer and query tokens using a standard regression objective, enabling it to proactively derive the expected 3D layout from high-level semantic cues. The loss is computed over the valid image regions $M$ as the pixel-wise distance between the prediction and the pseudo-ground-truth:
\begin{equation}
    \mathcal{L}_{depth} = \frac{1}{|M|} \sum_{i \in M} \lVert D_{gt}^{(i)} - D_{pred}^{(i)} \rVert_2
\end{equation}
Through this geometric-aware pre-training, the Spatial Transformer learns to ground abstract semantic cues into accurate 3D spatial arrangements, providing a structural scaffold to guide the subsequent diffusion process for coherent image generation.
% --- Table 4: Text-to-Image ---
\begin{table*}[t]
\centering

\begin{minipage}{0.9\textwidth}
\caption{Evaluation results on the GenSpace benchmark for spatially-aware text-to-image generation. The best results are highlighted in \textbf{bold} and the second-best results are \underline{underlined}.}
\label{tab:genspace_t2i}
\setlength{\tabcolsep}{2pt}
\begin{tabularx}{\linewidth}{@{} l *{12}{C} c @{}}
\toprule
\multirow{2.5}{*}{Model} & \multicolumn{4}{c}{\textbf{Spatial Pose}} & \multicolumn{4}{c}{\textbf{Spatial Relation}} & \multicolumn{4}{c}{\textbf{Spatial Measurement}} & \multirow{2.5}{*}{\specialcell{\textbf{Avg.}\\ \textbf{Score}}} \\ 
\cmidrule(lr){2-5} \cmidrule(lr){6-9} \cmidrule(lr){10-13}
 & Camera & Object & Complex & \textit{Avg.} & Ego. & Allo. & Intri. & \textit{Avg.} & Size & ObjDis & CamDis & \textit{Avg.} & \\ \midrule
\multicolumn{14}{c}{\textit{Expertise Generative Model}} \\ \midrule

SD-XL \cite{podell2023sdxl} & 33.66 & 25.03 & 9.52 & 22.74 & 46.15 & 16.38 & 8.87 & 23.80 & 23.89 & 33.76 & 22.75 & 26.80 & 24.45 \\
DALL-E 3 \cite{betker2023improving} & 50.37 & 46.81 & 10.92 & 36.03 & 65.74 & 17.45 & 16.63 & 33.27 & 30.32 & \textbf{41.91} & 25.69 & 32.64 & 34.03 \\
SD-3.5-L \cite{esser2024scaling} & 42.85 & 31.48 & 5.90 & 26.74 & 73.03 & 11.15 & \underline{23.55} & 35.91 & 31.03 & 33.05 & 24.83 & 29.64 & 30.76 \\
FLUX.1-dev \cite{flux2024} & 40.42 & 31.11 & 12.28 & 27.94 & 63.39 & 13.17 & 19.40 & 31.99 & 29.16 & 30.72 & 31.98 & 30.62 & 30.18 \\
Qwen-Image \cite{wu2025qwen} & 54.59 & 49.96 & 19.89 & 41.48 & 63.83 & 10.04 & 20.17 &  31.34 & 25.84 & 33.24 & 20.22 & 26.43 & 33.09 \\
Seedream-3.0 \cite{gao2025seedream} & 53.75 & 61.62 & 13.70 & 43.02 & 84.84 & 18.56 & 17.02 & 40.14 & 26.24 & 30.89 & 26.13 & 27.75 & 36.97 \\
\midrule
\multicolumn{14}{c}{\textit{Unified Generative Model}} \\ \midrule
UniWorld-V1 \cite{lin2025uniworld} & 23.72 & 24.59 & 15.78 & 21.36 & 59.62 & 17.09 & 13.59 & 30.10 & 31.74 & 18.22 & 30.85 & 26.94 & 26.13 \\
BAGEL \cite{deng2025emerging} & 43.34 & 46.65 & 13.47 & 34.49 & 72.10 & \underline{22.53} & 19.12 & 37.92 & 30.77 & 36.86 & 29.01 & 32.21 & 34.87 \\
Gemini-2.0-Flash \cite{google2025gemini} & 54.77 & 52.93 & 10.92 & 39.54 & 81.85 & 17.50 & 14.07 & 37.81 & 24.61 & 28.04 & 31.13 & 27.93 & 35.09 \\
GPT-4o \cite{openai2025gpt4o} & \underline{59.41} & \underline{62.72} & \textbf{25.01} & \underline{49.05} & \textbf{94.55} & 21.21 & 19.08 & \underline{44.95} & 30.47 & \underline{41.33} & \underline{35.19} & \underline{35.66} & \underline{43.22} \\ 
\midrule
OmniGen2 (base model)  & 40.44 & 41.96 & 12.82 & 31.74 & 61.88 & 17.65 & 14.46 & 31.33 & \underline{32.68} & 36.83 & 27.29 & 32.27 & 31.78 \\
\rowcolor[HTML]{E6F0FF} \textbf{SpatialFusion} & \textbf{73.10} & \textbf{71.30} & \underline{23.05} & \textbf{55.82} & \underline{84.93} & \textbf{26.01} & \textbf{25.02} & \textbf{45.32} & \textbf{35.75} & 39.47 & \textbf{38.32} & \textbf{37.85} & \textbf{46.33} \\
\bottomrule
\end{tabularx}
\end{minipage}
\end{table*}

% --- Table 5: Image Editing ---
\begin{table*}[t]
\centering
\small
\caption{Evaluation results on the GenSpace benchmark for spatially-aware instruction-based image editing.}
\label{tab:genspace_editing}
\setlength{\tabcolsep}{2pt}
\begin{tabularx}{0.9\textwidth}{@{} l *{12}{C} c @{}}
\toprule
\multirow{2.5}{*}{Model} & \multicolumn{4}{c}{\textbf{Spatial Pose}} & \multicolumn{4}{c}{\textbf{Spatial Relation}} & \multicolumn{4}{c}{\textbf{Spatial Measurement}} & \multirow{2.5}{*}{\specialcell{\textbf{Avg.}\\ \textbf{Score}}} \\ 
\cmidrule(lr){2-5} \cmidrule(lr){6-9} \cmidrule(lr){10-13}
 & Camera & Object & Complex & \textit{Avg.} & Ego. & Allo. & Intri. & \textit{Avg.} & Size & ObjDis & CamDis & \textit{Avg.} & \\ \midrule
\multicolumn{14}{c}{\textit{Expertise Generative Model}} \\ \midrule
Instruct-P2P \cite{brooks2023instructpix2pix} & 5.02 & 4.49 & 0.00 & 3.17 & 55.71 & \textbf{43.36} & 8.44 & 35.84 & 8.33 & 4.09 & 3.96 & 5.46 & 14.82 \\
ICEdit \cite{zhang2025enabling} & 4.04 & 5.61 & 0.23 & 3.29 & 63.36 & 42.40 & 12.52 & 39.43 & 9.37 & 5.35 & 5.46 & 6.73 & 16.48 \\
Step1X-Edit \cite{liu2025step1x} & 3.78 & 5.70 & 0.02 & 3.17 & 70.01 & 30.06 & 14.45 & 38.17 & \textbf{18.03} & 4.65 & 3.28 & 8.65 & 16.67 \\
SeedEdit \cite{shi2024seededit} & 23.51 & 16.03 & 0.78 & 13.44 & \underline{85.91} & 34.33 & \textbf{22.49} & \underline{47.58} & 11.46 & 7.03 & 8.80 & 9.10 & 23.37 \\
FLUX.1-Kontext-dev \cite{flux2025kontext} & 37.89 & 30.27 & 0.05 & 22.74 & 64.57 & 33.52 & 16.33 & 38.14 & 7.86 & 9.46 & 6.48 & 7.93 & 22.94 \\
Qwen-Image-Edit \cite{wu2025qwen} & 46.89 & 45.00 & 0.80 & 30.90 & 76.07 & 31.25 & 18.62 & 41.98  & 9.99 & 8.62 & 4.18 & 7.60  & 26.83  \\

\midrule
\multicolumn{14}{c}{\textit{Unified Generative Model}} \\ \midrule
UniWorld-V1 \cite{lin2025uniworld} & 11.59 & 16.86 & 0.00 & 9.48 & 68.36 & 20.99 & 19.05 & 36.13 & 9.31 & 4.84 & 4.19 & 6.11 & 17.24 \\

BAGEL \cite{deng2025emerging} & 45.37 & 49.55 & 0.77 & 31.90 & 78.51 & 38.74 & 17.03 & 44.76 & 11.11 & 6.79 & 4.94 & 7.61 & 28.09 \\

Gemini-2.0-Flash \cite{google2025gemini} & 46.81 & 38.12 & 0.17 & 28.37 & 81.19 & 33.88 & 18.50 & 44.52 & 7.02 & 5.04 & 8.63 & 6.90 & 26.60 \\
GPT-4o \cite{openai2025gpt4o} & \underline{54.38} & \underline{49.94} & \underline{1.80} & \underline{35.37} & \textbf{88.47} & 33.62 & 20.55 & 47.55 & 14.05 & \textbf{9.97} & \textbf{14.45} & \textbf{12.82} & \underline{31.91} \\ 
\midrule
OmniGen2 \cite{wu2025omnigen2} & 39.67 & 49.04 & 0.03 & 29.58 & 81.06 & 30.32 & 16.86 & 42.75 & 10.27 & 6.25 & 2.11 & 6.21 & 26.18  \\
\rowcolor[HTML]{E6F0FF} \textbf{SpatialFusion} & \textbf{64.80} & \textbf{72.72} & \textbf{2.50} & \textbf{46.67} & 82.14 & \underline{42.69} & \underline{19.47} & \textbf{48.10} & \underline{16.02} & \underline{9.44} & \underline{8.87} & \underline{11.44} & \textbf{35.40} \\
\bottomrule
\end{tabularx}
\end{table*}

\textbf{Stage 2: Geometry-Guided Joint Training}. 
With the Spatial Transformer pre-trained, we incorporate all tasks described in Sec. \ref{sec:3.2} and unfreeze the diffusion backbone to facilitate end-to-end joint optimization. The Spatial Transformer produces on-the-fly metric-depth predictions, which are directly injected into the diffusion backbone as explicit spatial conditioning signals to guide the generative denoising process.
To maintain high accuracy of intermediate 3D structures while simultaneously steering image synthesis, the architecture is optimized with a joint loss function:
\begin{equation}
    \mathcal{L}_{total} = \mathcal{L}_{diff} + \lambda \mathcal{L}_{depth}
\end{equation}
where $\mathcal{L}_{diff}$ represents the primary diffusion training objective, and $\mathcal{L}_{depth}$ is the explicit geometric regression loss strictly inherited from Stage 1. The hyperparameter $\lambda$ balances the generative quality and the rigid geometric structural constraints. 

Throughout both stages, we adopt a coarse-to-fine curriculum learning strategy. In each stage, the model is first trained on a large-scale data mixture to establish broad multimodal alignment and a robust geometric foundation, followed by fine-tuning on a curated high-quality subset to improve spatial precision and visual fidelity.
By combining this data-scaling curriculum with our two-stage progressive training paradigm, the model effectively endows the MLLM with intrinsic 3D geometric awareness while providing explicit geometric guidance to the diffusion model. As a result, the unified architecture internalizes rich spatial priors, achieving strong controllability and consistent structural alignment in both complex spatially-aware generation and geometry-grounded image editing.

\section{Experiments}

\subsection{Implementation Details}
We implement \textbf{SpatialFusion} based on the OmniGen2 \cite{wu2025omnigen2} backbone, which utilizes Qwen2.5-VL-3B \cite{bai2025qwen2} as the MLLM and a custom-trained diffusion decoder with approximately 4 billion parameters. To inject geometric awareness without significantly increasing model footprint, we introduce a lightweight parallel Spatial Transformer, comprising only 10 Transformer layers following the Qwen2.5-VL architecture. These layers are initialized from scratch to ensure specialized learning of spatial representations and share self-attention with the MLLM backbone through a uniform layer mapping strategy. 
To ensure robust optimization across large-scale and diverse data distributions, we utilize 16 NVIDIA H200 GPUs with a global batch size of 256. Our two-stage training strategy leverages data with diverse image aspect ratios via a bucket-based sampling scheme, facilitating stable optimization and improved generalization. More details are given in Appendix C.

\begin{table*}[t]
\centering
\small
\caption{Evaluation of text-to-image generation ability on GenEval, T2I-CompBench++, and DPG-Bench.}
\label{tab:t2i_results}
\setlength{\tabcolsep}{2pt} % 紧凑列间距
\resizebox{0.9\textwidth}{!}{
\begin{tabular}{@{} l *{7}{c} *{6}{c} c @{}}
\toprule
\multirow{2.5}{*}{Method} & \multicolumn{7}{c}{\textbf{GenEval}} & \multicolumn{6}{c}{\textbf{T2I-CompBench++}} & \textbf{DPG-Bench} \\ 
\cmidrule(lr){2-8} \cmidrule(lr){9-14} \cmidrule(l){15-15}
 & Overall & SingObj & TwoObj & Counting & Color & Pos. & ColorAttr & Color & Shape & Texture & 2D-Spa. & 3D-Spa. & Num. & Avg \\ \midrule
\multicolumn{15}{c}{\textit{Expertise Generative Model}} \\ \midrule
PixArt-alpha \cite{chen2023pixart} & 0.48 & 0.98 & 0.50 & 0.44 & 0.80 & 0.08 & 0.07 & 66.9 & 49.3 & 64.8 & 20.6 & 39.0 & 50.3 & 71.11 \\
SD-XL \cite{podell2023sdxl} & 0.55 & 0.98 & 0.74 & 0.39 & 0.85 & 0.15 & 0.23 & 58.8 & 46.9 & 53.0 & 21.3 & 35.7 & 49.9 & 79.26 \\
DALL-E 3 \cite{betker2023improving} & 0.67 & 0.96 & 0.87 & 0.47 & 0.83 & 0.43 & 0.45 & 77.9 & \textbf{62.1} & 70.4 & 28.7 & 37.4 & 59.3 & 83.50 \\
SD3 \cite{esser2024scaling} & 0.74 & 0.99 & 0.94 & 0.72 & 0.89 & 0.33 & 0.60 & \underline{81.3} & 58.9 & \underline{73.3} & 32.0 & 40.8 & 61.7 & 84.08 \\
FLUX.1-dev \cite{flux2024} & 0.66 & 0.98 & 0.79 & 0.73 & 0.77 & 0.22 & 0.45 & 74.1 & 57.2 & 69.2 & 28.6 & 38.7 & 61.9 & 83.79 \\ \midrule
\multicolumn{15}{c}{\textit{Unified Generative Model}} \\ \midrule
Show-o \cite{xie2024show} & 0.98 & 0.80 & 0.66 & 0.84 & 0.31 & 0.50 & 0.68 & - & - & - & - & - & -  & 67.27  \\
UniWorld-V1 \cite{lin2025uniworld} & 0.80 & 0.99 & 0.93 & \textbf{0.81} & 0.89 & 0.74 & \underline{0.71} & 61.8 & 33.5 & 47.4 & 27.5  & 40.5 & 55.3  & 81.38 \\
BAGEL \cite{hu2024ella} & \underline{0.82} & 0.99 & 0.94 & \textbf{0.81} & 0.88 & 0.64 & 0.63 & 81.0 & 56.2 & 70.8 & 35.4 & \underline{41.9} & \underline{64.7} & - \\

Janus-Pro \cite{chen2025janus} & 0.80 & 0.99 & 0.89 & 0.59 & \underline{0.90} & \textbf{0.79} & 0.66 & - & - & - & - & - & -  & \underline{84.17} \\
% GPT-4o \cite{openai2025gpt4o} & 0.85 & 0.99 & 0.92 & 0.85 & 0.91 & 0.75 & 0.66 & - & - & - & - & - & - & - \\ 
\midrule
OmniGen2 \cite{wu2025omnigen2} & 0.80 & \textbf{1} & \textbf{0.95} & 0.64 & 0.89 & 0.55 & \textbf{0.76} & 78.8 & 56.4 & 72.3 & \underline{38.8} & 41.4 & 64.4 & 83.57 \\
\rowcolor[HTML]{E6F0FF} \textbf{SpatialFusion} & \textbf{0.84} & \textbf{1} & \textbf{0.95} & \underline{0.78} & \textbf{0.92} & \underline{0.76} & \underline{0.71} & \textbf{81.5} & \underline{59.0} & \textbf{74.0} & \textbf{40.7} & \textbf{43.6} & \textbf{66.4} & \textbf{84.28} \\ \bottomrule
\end{tabular}
}
\end{table*}

\subsection{Main Results on GenSpace}

GenSpace \cite{wang2025genspace} serves as a specialized benchmark designed to comprehensively assess the spatial awareness of current image generation models, evaluating dimensions including Spatial Pose, Spatial Relations, and Spatial Measurement.
While the original study reveals significant limitations in existing models, our \textbf{SpatialFusion} framework delivers a marked improvement across all spatial dimensions, achieving the highest overall average scores in both generation and editing tasks, as shown in Table ~\ref{tab:genspace_t2i} and Table ~\ref{tab:genspace_editing}. 

\textbf{Text-to-Image Generation.}
% In the text-to-image task, \textbf{SpatialFusion} achieves an overall average score of 46.19, setting a new state-of-the-art. \textbf{1) Substantial Gains in Spatial Pose.}
\textbf{(1) In Spatial Pose}, 
while competitive unified models like GPT-4o plateau at roughly 60\% accuracy in Camera and Object Pose tasks, \textbf{SpatialFusion} significantly improves upon this, exceeding 70\% in both sub-categories, demonstrating a superior capability to effectively ground objects according to precise directional instructions. 
% \textbf{2) Progress in Challenging Spatial Relations and Measurement.}
\textbf{(2) In Spatial Relation}, most generative models heavily rely on egocentric priors and struggle with allocentric and intrinsic spatial understanding. In contrast, \textbf{SpatialFusion} shows notable improvement, achieving relative gains of approximately 23\% in allocentric tasks and 31\% in intrinsic relationship tasks over the strongest baseline (GPT-4o). This demonstrates a superior ability to accurately translate complex relational prompts into structurally coherent 2D scene layouts.
\textbf{(3) In Spatial Measurement}, a domain where nearly all models fail to generate images following specific quantitative constraints, our method achieves the highest overall average, proving its effectiveness in generating images that strictly adhere to precise quantitative constraints under the guidance of internally generated metric-depth maps.

\textbf{Instruction-based Image Editing.}
\textbf{(1) In Spatial Pose}, our method demonstrates strong structural versatility in pose manipulation—a task that inherently demands 3D orientation reasoning. While existing models struggle to modify object poses and shapes, our model achieves high accuracy in object pose manipulation, surpassing top unified baselines such as GPT-4o by a relative margin of over 45\%. This capability enables successful rotation or reorientation of objects according to instructions while strictly preserving their original textures and identities.
\textbf{(2) In Spatial Relation}, \textbf{SpatialFusion} effectively improves the ability to add new objects into established scenes according to complex spatial descriptions. It achieves the leading average score, underscoring its enhanced capacity to seamlessly insert objects at precise locations—whether defined relative to the viewer (egocentric) or to other scene entities (allocentric)—without introducing global structural degradation or triggering unintended regeneration of the original background.
\textbf{(3) In Spatial Measurement}, results indicate that precise quantitative editing—such as specifying the exact distance between existing objects—remains a profound bottleneck for current generative models. Nevertheless, \textbf{SpatialFusion} pushes the boundaries of the baseline model, unlocking the potential for executing metric-aware scale transformations directly from textual instructions.

% \textbf{SpatialFusion} addresses these shortcomings through two key advancements: \textbf{1) Structural Versatility}: Unlike specialized models that are tethered to the original image structure, our model can execute complex spatial edits—such as rotating an object's pose or altering its fundamental shape—based on natural language instructions. \textbf{2) Enhanced Identity Consistency}: A notable strength of our approach is the dramatic improvement in consistency before and after editing. While traditional unified models often discard the original image's details during a spatial edit, \textbf{SpatialFusion} preserves the texture, lighting, and identity of objects while precisely relocating or reorienting them. This ensures that the "edit" feels like a localized transformation rather than a global replacement.

\subsection{Main Results on Standard Benchmark}
To verify the versatility and robustness of \textbf{SpatialFusion}, we conduct extensive evaluations on several widely‑recognized benchmarks, spanning from text‑to‑image generation to image editing tasks, as presented in Table~\ref{tab:t2i_results} and Table~\ref{tab:editing_results}.

\textbf{Text-to-Image Generation.}
We evaluate \textbf{SpatialFusion} on three representative benchmarks: GenEval \cite{ghosh2023geneval}, T2I-CompBench++ \cite{huang2025t2i}, and DPG-Bench \cite{hu2024ella}. Compared to the backbone OmniGen2, our method achieves notable performance gains. We have the following observations:
\textbf{(1) Strong generalization across diverse generation tasks.}
Beyond spatial tasks, we observe that performance on other T2I dimensions (e.g., Color, Texture, and Shape) remains competitive or even improves across three diverse benchmarks. This suggests that internalized 3D geometric awareness does not merely serve as a spatial constraint but acts as a foundational prior that enhances overall generative quality.
\textbf{(2) Robustness across varying prompt complexities.} The consistent improvements across GenEval, T2I-CompBench++ (short, structured) and DPG-Bench (long, descriptive) underscore that \textbf{SpatialFusion} is a highly capable and efficient generator for diverse instructions, proving that the MoT architecture does not compromise the MLLM’s inherent linguistic flexibility. 
\textbf{(3) Generalization to compositional and spatial reasoning.} 
SpatialFusion significantly outperforms the baseline on the Position metric of GenEval, as well as on the 2D-spatial and 3D-spatial metrics of T2I-CompBench++. 
These gains indicate improved generalization to composition-intensive scenarios, where accurate spatial relationships are required, rather than improvements limited to specific task heuristics.
\begin{figure}[b]
  \centering
  \includegraphics[width=\linewidth]{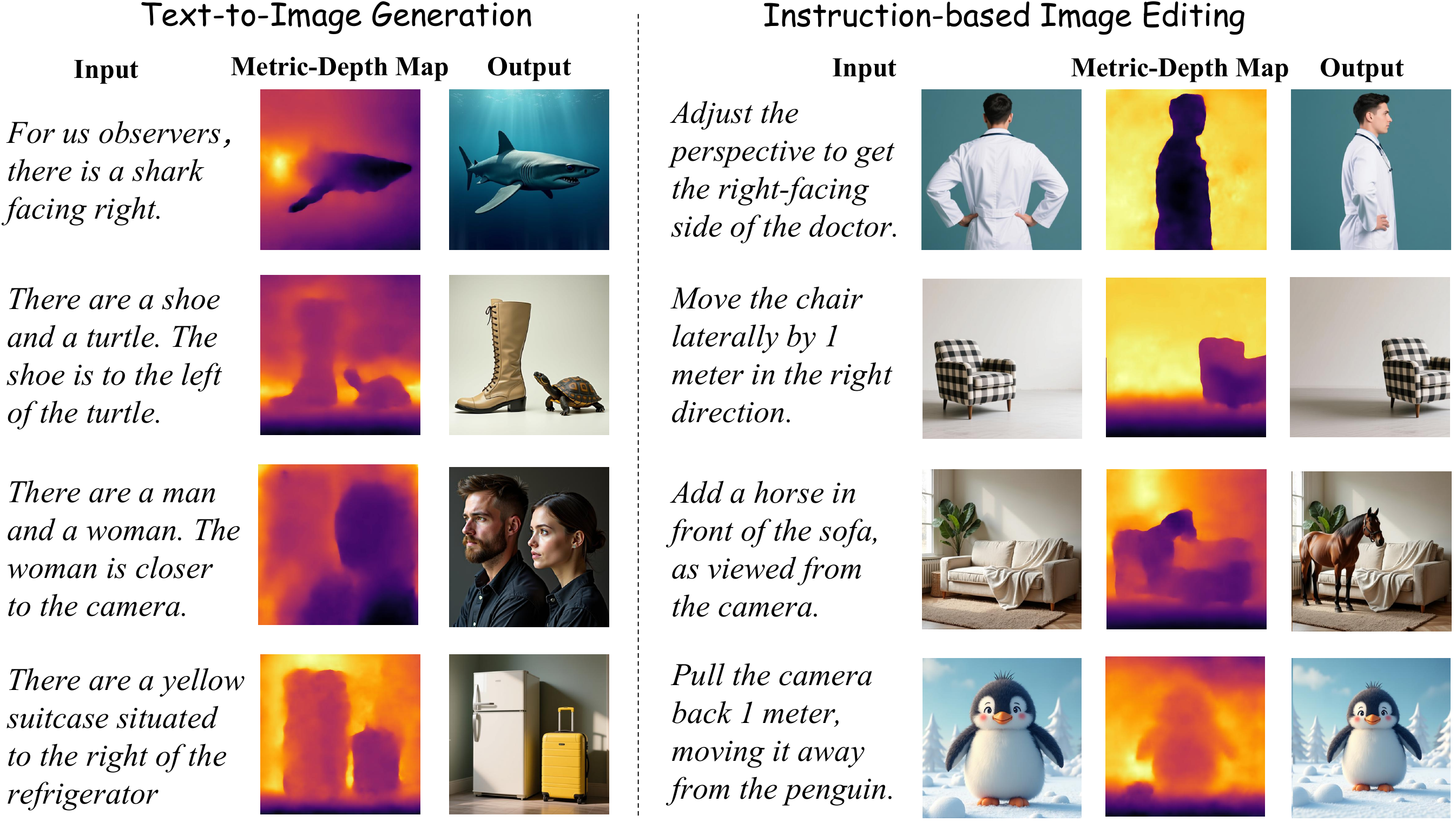}
  \caption{Qualitative results of image synthesis guided by derived intrinsic metric-depth maps.}
  \label{fig:vis1}
  \Description{}
\end{figure}

\begin{table*}[t]
\centering
\small
\caption{Evaluation of instruction-based image editing ability on ImgEdit-Bench and GEdit-Bench-EN.}
\label{tab:editing_results}
\setlength{\tabcolsep}{2pt} % 极限压缩间距以确保列名不换行
\resizebox{0.85\textwidth}{!}{
\begin{tabular}{@{} l *{9}{c} c ccc @{}}
\toprule
\multirow{2.5}{*}{Model} & \multicolumn{10}{c}{\textbf{ImgEdit-Bench}} & \multicolumn{3}{c}{\textbf{GEdit-Bench-EN}} \\ 
\cmidrule(lr){2-11} \cmidrule(lr){12-14}
 & Add & Adjust & Extract & Replace & Remove & Background & Style & Hybrid & Action & Overall& SC & PQ & O \\ \midrule
\multicolumn{14}{c}{\textit{Expertise Generative Model}} \\ \midrule

MagicBrush \cite{zhang2023magicbrush} & 2.84 & 1.58 & 1.51 & 1.97 & 1.58 & 1.75 & 2.38 & 1.62 & 1.22 & 1.90 & 4.68 & 5.66 & 4.52 \\
Instruct-P2P \cite{brooks2023instructpix2pix} & 2.45 & 1.83 & 1.44 & 2.01 & 1.50 & 1.44 & 3.55 & 1.20 & 1.46 & 1.88 & 3.58 & 5.49 & 3.68 \\
AnyEdit \cite{yu2025anyedit} & 3.18 & 2.95 & 1.88 & 2.47 & 2.23 & 2.24 & 2.85 & 1.56 & 2.65 & 2.45 & 3.18 & 5.82 & 3.21 \\
UltraEdit \cite{zhao2024ultraedit} & 3.44 & 2.81 & \underline{2.13} & 2.96 & 1.45 & 2.83 & 3.76 & 1.91 & 2.98 & 2.70 & - & - & - \\
Step1X-Edit \cite{liu2025step1x} & \underline{3.88} & 3.14 & 1.76 & 3.40 & 2.41 & 3.16 & 4.63 & 2.64 & 2.52 & 3.06 & 7.09 & 6.76 & \underline{6.70} \\
ICEdit \cite{zhang2025enabling} & 3.58 & 3.39 & 1.73 & 3.15 & 2.93 & 3.08 & 3.84 & 2.04 & 3.68 & 3.05 & 5.11 & 6.85 & 4.84 \\\midrule
\multicolumn{14}{c}{\textit{Unified Generative Model}} \\ \midrule
OmniGen \cite{xiao2025omnigen} & 3.47 & 3.04 & 1.71 & 2.94 & 2.43 & 3.21 & 4.19 & 2.24 & 3.38 & 2.96 & 5.96 & 5.89 & 5.06 \\
UniWorld-V1 \cite{lin2025uniworld} & 3.82 & \underline{3.64} & \textbf{2.27} & 3.47 & \underline{3.24} & 2.99 & 4.21 & \underline{2.96} & 2.74 & 3.26 & 4.93 & \underline{7.43} & 4.85 \\ 
BAGEL \cite{hu2024ella} & 3.56 & 3.31 & 1.70 & 3.30 & 2.62 & 3.24 & 4.49 & 2.38 & 4.17 & 3.20 & \underline{7.36} & 6.83 & 6.52 \\
Gemini-2.0-flash \cite{google2025gemini} & - & - & - & - & - & - & - & - & - & - & 6.73 & 6.61 & 6.32 \\
% GPT-4o \cite{openai2025gpt4o} & 4.61 & 4.33 & 2.90 & 4.35 & 3.66 & 4.57 & 4.93 & 3.96 & 4.89 & 4.20 & 7.85 & 7.62 & 6.52 \\ 
\midrule
OmniGen2 \cite{wu2025omnigen2} & 3.57 & 3.06 & 1.77 & \underline{3.74} & 3.20 & \underline{3.57} & \textbf{4.81} & 2.52 & \underline{4.68} & \underline{3.44} & 7.16 & 6.77 & 6.41 \\
\rowcolor[HTML]{E6F0FF} \textbf{SpatialFusion } & \textbf{4.17} & \textbf{3.89} & 1.80 & \textbf{4.63} & \textbf{4.41} & \textbf{4.11} & \underline{4.65} & \textbf{3.71} & \textbf{4.70} & \textbf{4.01} & \textbf{7.92} & \textbf{7.68} & \textbf{7.58} \\ \bottomrule
\end{tabular}
}
\end{table*}

\textbf{Instruction-based Image Editing.}
To rigorously assess editing capabilities, we evaluate our model on ImgEdit-Bench \cite{ye2025imgedit} and GEdit-Bench-EN \cite{liu2025step1x}. Results show that our model exhibits strong generalization across diverse editing scenarios, yielding the following advantages:
\textbf{(1) Improved generalization with enhanced semantic and perceptual fidelity.}
On GEdit-Bench, we observe consistent gains in both Semantic Consistency (SC) and Perceptual Quality (PQ). On ImgEdit-Bench, our method achieves substantial improvements across diverse editing tasks. This indicates that our spatial guidance does not merely enforce "hard" geometric constraints but also harmonizes with high-level semantic intent without degrading synthesis quality.
\textbf{(2) Robust performance on structure-sensitive editing tasks.}
Our model significantly outperforms the baseline on challenging editing scenarios in ImgEdit-Bench, achieving relative improvements of 17\% (object addition), 24\% (replacement), and 38\% (removal), demonstrating that explicit geometric guidance leads to stronger adaptability in accurate object placement and coherent background completion.
\textbf{(3) Consistent editing across benchmarks.}
Our model achieves consistent performance gains across all benchmarks, where aggregate scores reflect both accurate editing and strong content preservation. This demonstrates improved robustness and stability across diverse editing operations, highlighting its generalization ability.
\subsection{Qualitative Comparisons}
\begin{figure*}[t]
  \centering
  \includegraphics[width=\textwidth]{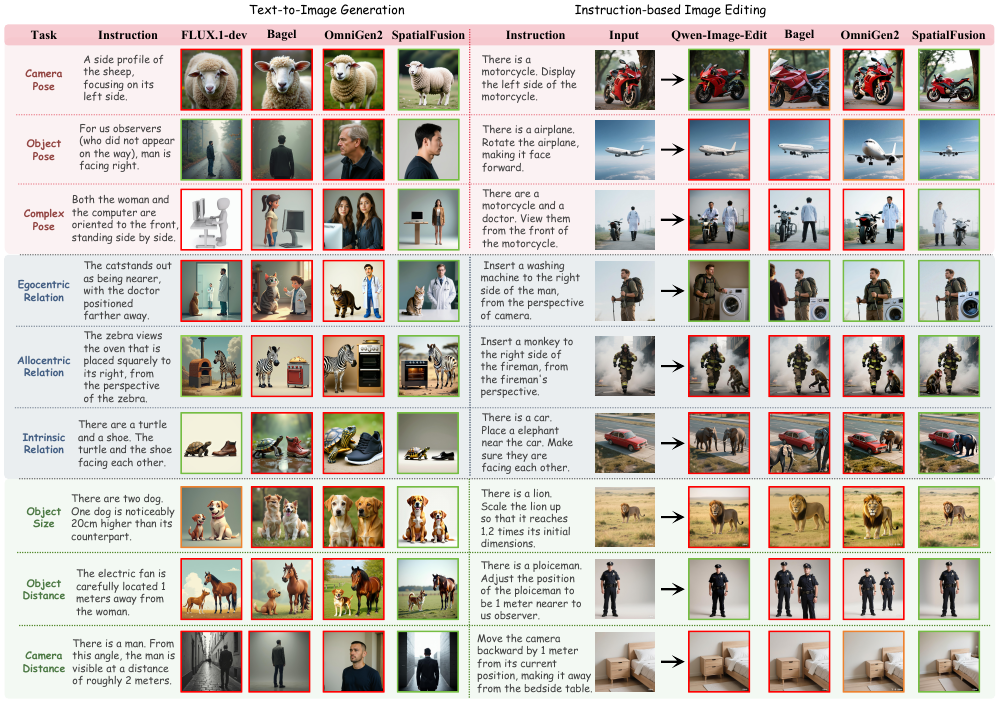}
  \caption{Qualitative comparison of spatially-aware image generation. Our method achieves the best performance.}
  \label{fig:vis2}
  \Description{}
\end{figure*}

In Fig.~\ref{fig:vis1}, we present qualitative examples of internally derived metric-depth maps alongside the corresponding generated images. Our method employs a parallel spatial transformer to derive metric-depth maps of the target scene, which then provide rigorous geometric guidance for subsequent image generation and editing. In Fig. ~\ref{fig:vis2}, we compare our method with several baseline models on spatially-aware generation tasks. While baselines suffer from various spatial errors, \textbf{SpatialFusion} consistently respects object poses, spatial layouts, and metric constraints.
\begin{figure}[h]
    \centering
    % 第一张子图
    \begin{subfigure}{0.42\linewidth}
        \centering
        \includegraphics[width=\textwidth]{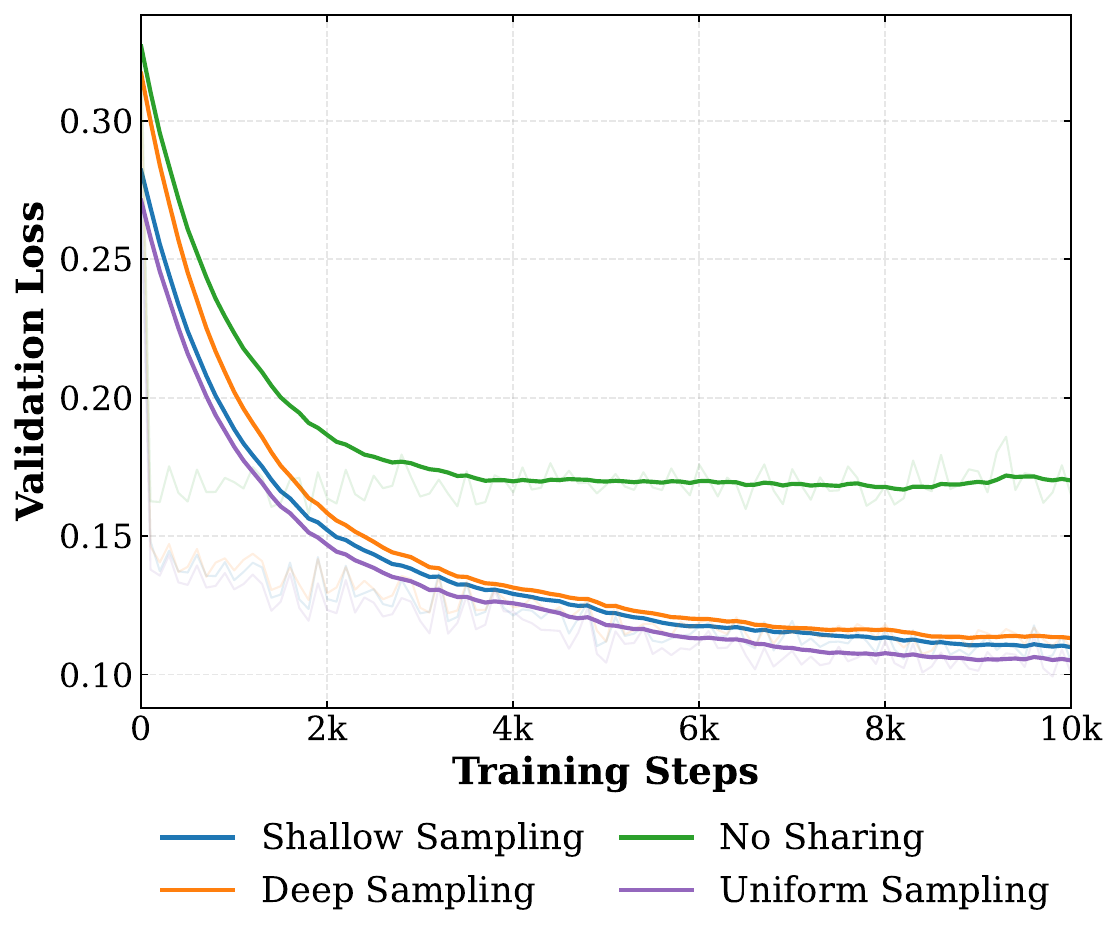}
        \caption{} % <--- 必须有这个，才会显示 (a)
        \label{fig:ablation1}
    \end{subfigure}%
    \hfill 
    % 第二张子图
    \begin{subfigure}{0.5\linewidth}
        \centering
        \includegraphics[width=\textwidth]{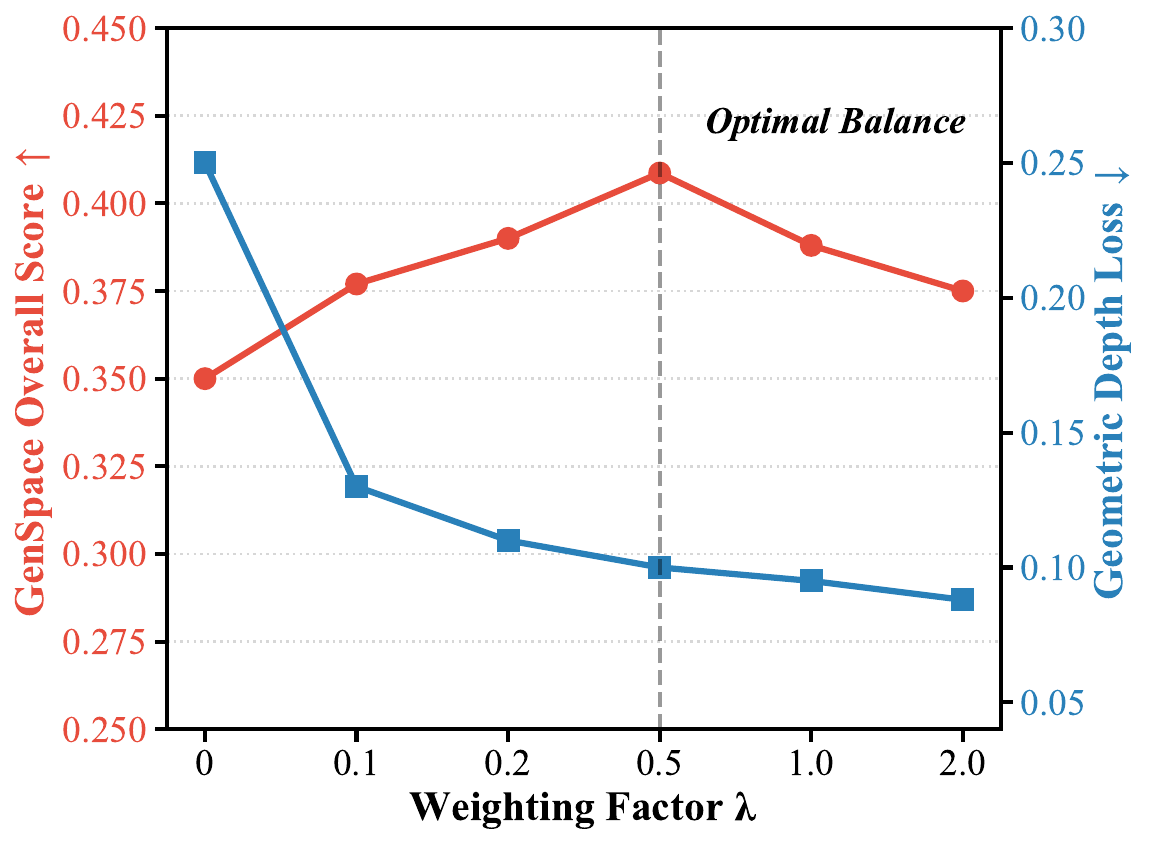}
        \caption{} % <--- 必须有这个，才会显示 (b)
        \label{fig:ablation2}
    \end{subfigure}
    
\caption{(a) Ablation study on the shared attention sampling mechanism; (b) Ablation study on the weighting factor $\lambda$.}
    \label{fig:ablation_study}
\end{figure}

\subsection{In-Depth Analysis}

\textbf{Effect of Shared Attention on Geometry Prediction.} To evaluate shared attention, we fine-tune the model on a spatially-aware subset in Stage 1, comparing different strategies for selecting which MLLM layers participate in self-attention  with the 10 Spatial Transformer layers. As shown in Fig.~\ref{fig:ablation1}, we compare their geometry prediction performance based on the average validation depth loss. Notably, all shared-attention variants consistently outperform the independent baseline (\textit{No Sharing}), highlighting the importance of semantic priors in deriving target geocentric structure. Among them, \textit{Uniform Sampling} across MLLM layers performs best, followed by \textit{Shallow Sampling} (first 10 layers) and \textit{Deep Sampling} (last 10 layers). This suggests that interacting with a balanced spectrum of semantic features—spanning from low-level to high-level—improves geometric reasoning and metric-depth derivation.

\textbf{Effect of Geometric Guidance on Generation Performance.} 
To examine the contribution of geometric guidance, we compare three configurations in the Table ~\ref{tab:ablation_and_efficiency}: \textit{w/o Injection} feeds no 3D information to the diffusion model, serving as a data-driven baseline, while \textit{Concatenation} and \textit{Addition} incorporate metric-depth features via channel-wise concatenation and element-wise addition to latent features, respectively.
Results on GenSpace reveal two key findings: (1) \textit{w/o Injection} yields marginal gains over base model, whereas incorporating 3D guidance (\textit{Concatenation} and \textit{Addition}) significantly improves performance, indicating that the model performance gains arise from intrinsic
3D geometric awareness rather than overfitting. (2) \textit{Addition} consistently outperforms \textit{Concatenation}. This is likely because concatenation changes the input dimensionality, disrupting pretrained priors, whereas addition acts as a residual that injects geometric constraints with minimal distribution shift.

\textbf{Analysis of Geometric Supervision Weight on Joint Training.} During Stage 2, the overall loss is a weighted sum of the image reconstruction loss ($\mathcal{L}_{diff}$) and the geometric auxiliary loss ($\mathcal{L}_{depth}$). Fig.~\ref{fig:ablation2} illustrates the model's overall performance on GenSpace alongside its metric-depth predictive capability (evaluated via the average validation depth loss) as the weighting factor $\lambda$ varies. The empirical trends reveal a trade-off: increasing $\lambda$ continuously strengthens the model's ability to derive metric-depth maps. However, an overly large $\lambda$ forces the network to over-focus on geometric derivation, which inadvertently weakens the diffusion model's primary image synthesis capabilities. Setting $\lambda = \text{0.5}$ achieves the optimal balance between geometric consistency and visual fidelity.

\begin{table}[h]
\centering
\caption{Ablation study on geometric guidance mechanisms and analysis of computational overhead. Latency is evaluated on a single NVIDIA H200 GPU averaged over 300 runs.}
\label{tab:ablation_and_efficiency}
\resizebox{\linewidth}{!}{
\begin{tabular}{lccccc}
\toprule
\multirow{2}{*}{Configuration} & \multicolumn{2}{c}{GenSpace (Avg. Score)}  & \multicolumn{2}{c}{Latency (s / image)} \\
\cmidrule(lr){2-3} \cmidrule(lr){4-5}
& T2I & Editing & T2I & Editing \\
\midrule
OmniGen2 (Base) & 31.78 & 26.18  & 12.32 & 30.24 \\
\textit{w/o Injection} & 35.22  & 29.80  & - & - \\
\textit{Concatenation} & 42.81 & 32.76  & - & - \\
\textit{Addition} (\textbf{SpatialFusion}) & \textbf{46.19} & \textbf{35.40}  & 12.58 {\small (+0.26)} & 31.07 {\small (+0.83)} \\
\bottomrule
\end{tabular}
}
\end{table}

\textbf{Analysis of Computational Overhead.} 
We evaluate the practical efficiency of our method in Table~\ref{tab:ablation_and_efficiency}. While introducing the lightweight spatial transformer, depth adapter, and VGGT DPT head introduces a modest 18.5\% increase in the total parameter count, the impact on inference latency is remarkably minimal. Our method incurs only a marginal time overhead—adding less than a 3\% delay per image for both text-to-image generation and image editing tasks. This slight computational cost is highly acceptable given the substantial leaps in spatial accuracy and editing consistency.

\section{Conclusion}
In this paper, we introduce \textbf{SpatialFusion}, a novel framework that seamlessly integrates 3D geometric awareness into unified generative models. By employing a MoT architecture, our method derives explicit metric-depth maps of target scenes, which are subsequently integrated into the diffusion model via a depth adapter to provide precise geometric constraints. Driven by a two-stage training strategy, \textbf{SpatialFusion} successfully internalizes accurate 3D spatial awareness while preserving pre-trained semantic priors. Experimental results demonstrate that our approach achieves superior geometric consistency and spatial fidelity in both text-to-image generation and image editing, marking a significant step toward spatial reasoning in generative systems.
\bibliographystyle{ACM-Reference-Format}
\bibliography{sample-base}

\end{document}